# Feeder Load Balancing Using Neural Network

Abhisek Ukil, Willy Siti, and Jaco Jordaan

Tshwane University of Technology, Private Bag X680,
Pretoria, 0001, South Africa
{abhiukil, willysiti, jakop_s2003}@yahoo.com

**Abstract.** The distribution system problems, such as planning, loss minimization, and energy restoration, usually involve the phase balancing or network reconfiguration procedures. The determination of an optimal phase balance is, in general, a combinatorial optimization problem. This paper proposes optimal reconfiguration of the phase balancing using the neural network, to switch on and off the different switches, allowing the three phases supply by the transformer to the end-users to be balanced. This paper presents the application examples of the proposed method using the real and simulated test data.

## 1 Introduction

Phase balancing in the distribution system have different needs, from minimizing the losses in the system to relieving the transformer during the peak time and so forth. There are a number of normally closed and normally opened switches in a distribution system. By changing the open/close status of the feeder switches, load currents can be transferred from feeder to feeder, that is, from the heavily loaded to the less loaded feeders. In South Africa, to reduce the unbalance current in a feeder the connection phases of some feeders are changed manually after some field measurement and software analysis. This is, however, time-consuming and unsuccessful many times [1].

With the uses of the artificial intelligence, telecommunication and power electronics equipments in the power system, it is becoming easier to automate the phase balancing problem. The automation implementation will be technically advantageous as well as economical for the utilities and the customers, in terms of the variable costs reduction and better service quality, respectively.

The approach proposed here uses the neural network which will be able to switch on/off the different switches and keep the phases balanced. Each load will cater only one of the three phases following the constraint that for each load only one switch (to the phase) should be closed, while other two should remain open. For each loading condition, the neural network will be trained for the relevant minimum loss configuration. This can be applied to the small networks, for example, six to fifteen houses as the unbalanced loads.

## 2 Problem Description

To balance the three phase currents in every segment and then depressing the neutral line current is a very difficult task. Using the manual trial and error technique, used most of the time in South Africa for phase balancing, interruption of the service continuity is unavoidable when changing the connection phases of the distribution transformers to the feeder [1].

In South Africa, a distribution feeder is usually a three-phase, four-wire system. It can be radial or open loop structure [3]. The example feeder shown in Fig. 1 has three phase conductors for the section between the main transformer and the different load points. We limit our present study to six load points, as shown in Fig. 1. To improve the system phase voltage and current unbalances, the connections between the specific feeders and the distribution transformers should be suitably arranged.

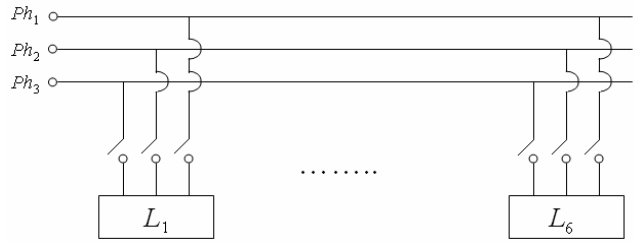

**Fig. 1.** Example Distribution Feeder

## 3 Feeder Reconfiguration Technique

In the case of a distribution system with some of the branches overloaded, and other branches lightly loaded, there is the need to reconfigure the networks such that loads are transferred from the heavily loaded feeder or transformers to the less loaded feeder (or transformers). The maximum load current the feeder conductor can take may be taken as the reference. The transfer of load must be done by satisfying the predefined objective to have minimum real power loss. Consequently, network reconfiguration may be redefined as the rearrangement of the network such as to minimize the total real power losses arising from the line branches. Mathematically, the total power loss can be expressed as follows [2]:

$$\sum_{i=1}^{n} r_i \frac{P_i^2 + Q_i^2}{|V_i|^2} ,  \quad (1)$$

where, $r_i$, $P_i$, $Q_i$, $V_i$ are respectively the resistance, the active power, the reactive power and the voltage of the branch $i$, and $n$ is the total number of branches in the system. Due to some practical considerations, there could be a constraint on the number of switch–on and off. Given a distribution system as shown in Fig. 1, a network with three phases with a known structure, the problem consists of finding a condition of balancing. The mathematical model [1] can be expressed as:

$$\mathbf{I}_{ph1k} = \sum_{i=1}^{3} sw_{k1i} \mathbf{I}_{ki} + \mathbf{I}_{ph1(k+1)} \quad (2)$$

$$\mathbf{I}_{ph2k} = \sum_{i=1}^{3} sw_{k2i}\mathbf{I}_{ki} + \mathbf{I}_{ph2(k+1)} \tag{3}$$

$$\mathbf{I}_{ph3k} = \sum_{i=1}^{3} sw_{k3i}\mathbf{I}_{ki} + \mathbf{I}_{ph3(k+1)} , \tag{4}$$

where, $\mathbf{I}_{ph1k}$, $\mathbf{I}_{ph2k}$ and $\mathbf{I}_{ph3k}$ represent the currents (phasors) per phase (1, 2 & 3) after the *k* point of connection; $sw_{k11},\ldots,sw_{k33}$ are different switches (the value of '1' means the switch is closed and '0' means it is open). Following the constraint of allowing only one breaker in each of the equations (2) – (4) to be closed, we can write the following set of modified constraints:

$$\sum_{i=1}^{3} sw_{k1i} - 1 = 0 \tag{5}$$

$$\sum_{i=1}^{3} sw_{k2i} - 1 = 0 \tag{6}$$

$$\sum_{i=1}^{3} sw_{k3i} - 1 = 0 . \tag{7}$$

## 4 Neural Network-Based Approach

In the proposed strategy in this paper, the neural network must control the switch-closing sequence of each load for the minimum power loss which will lead to the optimal phase balance. The inputs to the neural network are the unbalanced load currents (six in the current study) and the outputs are the switch closing sequences for each load.

The input layer of the network has *N* input neurons, *N* being number of unbalanced load currents to be controlled. The following column vector has been assumed as the input

$$C_{Sw} = [I_{L1}\ldots\ldots\ldots\ldots I_{LN}] . \tag{8}$$

The output of the network is in the range {1, 2, 3} for each load, i.e., which switch (to the specific phase) should be closed for that specific load.

### 4.1 Neural Network Structure

For this application, we used the radial basis network [3]. Experimentations with the backpropagation and the radial basis network indicated faster training and better convergence for the latter. Radial basis networks may require more neurons than the standard feed-forward backpropagation networks, but often they can be designed in a fraction of the time needed to train the standard feed-forward networks. They work best when many training vectors are available [4]. Matlab® neural network toolbox [5] has been used for the implementation. We experimented with different kinds of radial basis networks, but generalized regression neural network ("GRNN") [5] produced the best result. A generalized regression neural network is often used for function approximation. It has a radial basis layer and a special linear layer [5].

### 4.2 Network Training

We have used the neural network-based operation for the test data in following structure: real and simulated data for six loads.

The real data set consisted of unbalanced load data from a South African city. The test data set had average load current values per houses in a specific locality of the city for the different times of each day in a month. We randomly selected six houses as our test data for each specific time, and we tested our result on 500 data. Simulated data were generated using the computer following the real load data structure.

First, we used the Matlab®-based fast heuristic method [1] for balancing the unbalanced load data. Details of the algorithm can be referred to in [1], but we explain the necessary part briefly below.

We consider the loads to be equally distributed per phase, i.e., we assume two loads to be connected per phase. So, the problem is to find the optimum three sets of two loads, with *minimum* differences among the individual sums of the three sets. To achieve this, first we calculate the ideal phase balance current value $I_{ideal}$, which is equal to the one-third of the sum of the all six load currents $I_L$.

$$I_{ideal} = \frac{1}{3}\sum_{j=1}^{6} I_{L_j} . \tag{9}$$

In the second step, we optimally select our 3 sets of currents for the three phase currents $I_{ph}$, each set comprising of two load currents $\{I_j, j=1,2\}$.

$$I_{Load} = \{I_{L_j}, j=1,...,6\} , \tag{10}$$

$$I_{ph} = \{I_j, j=1,2\} \quad \text{where} \quad I_j \in I_{Load} . \tag{11}$$

Difference between the individual sum of these sets and the $I_{ideal}$ should be *minimum*, ideally 0 for the perfect phase balance. So, we need to find three sets of $\{I_j, j=1,2\}$, subject to the constraint,

$$\min \left| \sum_{j=1}^{2} I_j - I_{ideal} \right| , \quad \text{where} \quad I_j \in I_{Load} . \tag{12}$$

Following this, we obtain the output switching sequences as the target data set for training the networks. Using the output switching sequences and the input load currents, we calculate the balanced phase currents $I_{ph1}$, $I_{ph2}$ and $I_{ph3}$. For example, $I_{ph1}$ is calculated by adding the two load currents corresponding to the output switching sequences marked 1. Then we calculate the differences between $I_{ph1}$, $I_{ph2}$ and $I_{ph3}$, which ideally should be zero. The differences indicate the quality of the phase balance [1].

Using the real and simulated unbalanced load as the input vector, and the output switching sequences as the target vector, we trained the above-mentioned neural network. Then, we tested the network with different unbalanced load data set. The output was the optimal switching sequences of {1, 2, 3} for the three-phases as explained above. Using the similar procedure as explained above, we computed the balanced phase currents and the differences between the phase currents, which indicate the quality of the balance.

## 5 Application Results

An Intel® Celeron® 1.9 GHz, 256 MB RAM computer was used for the test. Test results of the neural network-based approach for the simulated six load data format are shown in Table 1 to 4, for three different sample data. Table 1 shows the unbalanced load (current) data, Table 2 the output switching sequences, Table 3 the balanced phase currents and Table 4 the absolute differences between the balanced phase currents. In Table 2 to 4, 'NN' is the abbreviation for the Neural Network-based approach and 'HEU' is the abbreviation for the Heuristic Method [1]-based approach.

**Table 1.** Unbalanced Load (Current) Data

| Data $I_L$ (A) | 1 | 2 | 3 |
|---|---|---|---|
| 1 | 89 | 35 | 45 |
| 2 | 85 | 0 | 67 |
| 3 | 74 | 90 | 87 |
| 4 | 38 | 21 | 64 |
| 5 | 56 | 87 | 30 |
| 6 | 45 | 112 | 90 |

**Table 2.** Output Switching Sequences

| Switch Seq. for 6 Loads | Data 1 | | Data 2 | | Data 3 | |
|---|---|---|---|---|---|---|
| | NN | HEU | NN | HEU | NN | HEU |
| 1 | 1 | 1 | 1 | 1 | 1 | 1 |
| 2 | 2 | 2 | 2 | 3 | 2 | 2 |
| 3 | 3 | 3 | 1 | 2 | 3 | 1 |
| 4 | 1 | 1 | 3 | 1 | 2 | 2 |
| 5 | 3 | 3 | 3 | 2 | 3 | 3 |
| 6 | 2 | 2 | 2 | 3 | 1 | 3 |

**Table 3.** Balanced Phase Currents

| $I_{Phase}$ (A) | Data 1 | | Data 2 | | Data 3 | |
|---|---|---|---|---|---|---|
| | NN | HEU | NN | HEU | NN | HEU |
| Phase 1 | 127 | 127 | 125 | 56 | 135 | 132 |
| Phase 2 | 130 | 130 | 112 | 177 | 131 | 131 |
| Phase 3 | 130 | 130 | 108 | 112 | 117 | 120 |

**Table 4.** Differences between Phase Currents

| Difference (A) | Data 1 | | Data 2 | | Data 3 | |
|---|---|---|---|---|---|---|
| | NN | HEU | NN | HEU | NN | HEU |
| Phase 1-2 | 3 | 3 | 13 | 121 | 4 | 1 |
| Phase 2-3 | 0 | 0 | 4 | 65 | 14 | 11 |
| Phase 3-1 | 3 | 3 | 17 | 56 | 18 | 12 |

### 5.1 Comments on Application Results

- Summary of the neural network-based approach in comparison with the heuristic method [1]-based one is as follows.
    - Neural Network performs better than Heuristic Method: 14%
    - Neural Network performs same as Heuristic Method: 67%

- o Neural Network performs worse than Heuristic Method: 10%
- o Neural Network fails to converge or gives erroneous result: 9%

- From the above summary, it should be noted that the neural network-based approach mostly works similar to the heuristic method [1]-based approach. Deviation of the results in the 10% worse cases are not that severe, however, the 14% better performance is a significant improvement, as the heuristic method proved to be the most efficient [1].
- Speed of operation (average computation time 0.14 seconds) is similar with the heuristic method [1], once the network is trained. For this reason, once the network is suitably trained, we save and use it as a neural network object.
- This approach can be extended to any number of unbalanced load data. But as the training data construction depends on the heuristic method [1], at this stage, we have to limit our studies to the number of load data exactly divisible by 3 so that the loads can be equally distributed per phase.

## 6 Conclusion

The neural network-based approach for phase balancing for a small size network (six load data) proved the feasibility of the proposed control. The result achieved from the Matlab® implementation pertain to that obtained using the heuristic method. Besides, neural network-based approach gives better result in 14% cases, which is a significant improvement. This encourages the implementation of the neural network-based strategy on a large size network.

## References

1. Ukil, A., Siti, M. and Jordaan, J.: MATLAB®-based Fast Load Balancing of Distribution System Using Heuristic Method. SAIEE Transactions (under review)
2. Chen, C.S. and Cho, M.Y.: Energy Loss Reduction by Critical Switches, of Distribution Feeders for Loss Minimization. IEEE Trans. Power Delivery 4 (1992)
3. Wasserman, P.D.: Advanced Methods in Neural Computing. Van Nostrand Reinhold, New York (1993)
4. Chen, S., Cowan, C.F.N. and Grant, P.M.: Orthogonal Least Squares Learning Algorithm for Radial Basis Function Networks. IEEE Trans. Neural Networks 2 (1991) 302-309
5. MATLAB® Documentation – Neural Network Toolbox, Version 6.5.0.180913a Release 13, Mathworks Inc., Natick, MA (2002)